\pdfoutput=1
\documentclass[11pt]{article}

\usepackage{ACL2023}

\usepackage{times}
\usepackage{latexsym}
\usepackage[T1]{fontenc}
\usepackage[utf8]{inputenc}
\usepackage{microtype}
\usepackage{inconsolata}
\usepackage{enumitem,kantlipsum}
\usepackage{graphicx}
\usepackage{caption}
\usepackage{amsmath}
\usepackage{amssymb}
\usepackage{pgfplots}
\pgfplotsset{compat=newest}
\usepackage{comment}
\usepackage{booktabs}
\usepackage{hyperref}
\usepackage[overload]{empheq}
\usepackage{makecell}
\usepackage{xcolor,colortbl}
\usepackage[ruled,vlined]{algorithm2e}
\usepackage{array,multirow}
\usepackage{comment}
\usepackage{pifont}
\usepackage{siunitx}
\usepackage{arydshln}
\usepackage{url}

\usepackage{newpxmath}
\setlist[itemize]{leftmargin=*}

\newcommand{\cmark}{\ding{51}}%
\newcommand{\xmark}{\ding{55}}%

\newcommand*{\myfont}{\fontfamily{qpl}\selectfont}
\DeclareTextFontCommand{\textmyfont}{\myfont}

\title{That's Optional: A Contemporary Exploration of "that" Omission in English Subordinate Clauses}

\author{Ella Rabinovich \\
The Academic College of Tel Aviv-Yaffo, Israel \\
\texttt{ellara@mta.ac.il}
}

\begin{document}
\maketitle
\begin{abstract}
The Uniform Information Density (UID) hypothesis posits that speakers optimize the communicative properties of their utterances by avoiding spikes in information, thereby maintaining a relatively uniform information profile over time. This paper investigates the impact of UID principles on syntactic reduction, specifically focusing on the optional omission of the connector "that" in English subordinate clauses. Building upon previous research, we extend our investigation to a larger corpus of written English, utilize contemporary large language models (LLMs) and extend the information-uniformity principles by the notion of entropy, to estimate the UID manifestations in the usecase of syntactic reduction choices.

\end{abstract}

\section{Introduction}
\label{sec:introduction}

%\ella{we can write this section later, once the full picture and the way we position the paper is clear; for this short paper it should include related work}

Exploiting the expressive richness of languages, speakers often convey the same messages in multiple ways. A body of research on \textit{uniform information density} (UID) puts forward the hypothesis that speakers tend to optimize the communicative effectiveness of their utterances when faced with multiple options for structuring a message. The UID hypothesis \citep{frank2008speaking, collins2014information, hahn2020universals} suggests that speakers tend to spread information evenly throughout an utterance, avoiding large fluctuations in the per-unit information content of an utterance, thereby decreasing the processing load on the listener.

The UID hypothesis has been used as an explanatory principle for phonetic duration \citep{bell2003effects, aylett2006language}, the choice between short- and long-form of words that can be used interchangeably, such as "info" and "information" \citep{mahowald2013info}, and word order patterns \citep{genzel2002entropy, maurits2010some, meister2021revisiting, tacl_a_00589}. Our work studies how UID principles affect the phenomenon of syntactic reduction -- the situation where a speaker has the choice of whether marking a subordinate clause in sentence with an optional subordinate conjunction (SCONJ) "that" or leave it unmarked, as in "My daughter mentioned [that] he looked good". The only study that tested the UID hypothesis computationally in the context of syntactic reduction is \citet{jaeger2006speakers}, followed by \citet{jaeger2010redundancy}, who studied the effect of multiple factors on the speaker choice of \textit{explicit} or \textit{implicit} "that" conjunction. Investigating sentences with main clause (MC, e.g., "My daughter mentioned") and subordinate clause (SC, e.g., "[that] he looked good"), connected by the optional SCONJ, the authors found that UID optimization was the most prominent factor affecting a speaker choice of "that" omission. Specifically, \citet{jaeger2010redundancy} investigated 6700 sentences extracted from the SwitchBoard spoken English dataset, and operationalized the UID principle by computing the surprisal (non-predictability) of the SC opening word (SC onset) using a statistical bigram language model computed from the corpus itself.

Our work studies the role of UID principle in syntactic reduction in multiple differing ways. First, we extend the investigation to a much larger corpus of informal \textit{written} English collected from social media. Second, we use contemporary large language models (LLMs) to estimate the operationalizations of information uniformity in syntactic reduction, suggesting the robustness of our findings. Finally, inspired by the information-theoretic nature of UID and prior art \citep{maurits2010some, meister2021revisiting}, we extend the SC onset surprisal UID manifestation with the notion of SC onset \textit{entropy} -- the information entropy of LLM distribution over SC opening word, conditioned on the main clause -- factor that turns out to have a complementary and significant effect. % on a writer decision on "that" omission.

The contribution of this work is, therefore, twofold: First, we collect and release a large and diverse corpus of nearly 100K sentences, where main and subordinate clauses are connected by the optional SCONJ "that".\footnote{All data and code are available at \url{https://github.com/ellarabi/uid-that-sc-omission}.} Second, we go above and beyond prior work by using transformer-based LLMs \citep{vaswani2017attention}, thereby providing a sound empirical evidence for UID principles associated with syntactic reduction decision, shedding a new and interesting light on the manifestation of UID in spontaneous written language.

  % and related work
%\input{chapters/2-related-work}
\section{Dataset}
\label{sec:dataset}

\subsection{Data Collection}

%\ella{collection of the reddit data -- the ample data for large-scale computational analysis; identification of the two sentence sets -- explicit and implicit usage of "that" via syntactic parsing; examples}

Our dataset in this work was collected from the \href{https://www.reddit.com/}{Reddit} discussions platform. Reddit is an online community-driven platform consisting of numerous forums for news aggregation, content rating, and discussions. %As of 2023, it had over ... million monthly active users, positioning it as the ... most popular social site in the US.\footnote{\url{https://thesmallbusinessblog.net/reddit-statistics/}} 
%Content entries are organized by areas of interest called subreddits, ranging from main forums that receive extensive attention to smaller ones that foster discussion on niche areas. 
Communication on discussion platforms often resembles a hybrid between speech and more formal writing, and findings from spoken language may extend to the spontaneous and informal style of social media. As such, Reddit data has been shown to exhibit code-switching patterns, similar to those found in spoken language \citep{rabinovich2019codeswitch}. We, therefore, believe that this data presents a good testbed for our analysis. % -- the optional usage of "that" SCONJ at the onset of a subordinate clause.

\paragraph{Data Extraction} 
We collected 2M posts and comments by over 20K distinct redditors spanning over 5K topical threads and years 2020--2022. We then split the data into sentences and filtered out sentences shorter than five or longer than 50 words. The remaining 487,614 sentences were parsed using the SOTA \href{https://github.com/nikitakit/self-attentive-parser}{benepar} syntactic parser, extracting two sentence types with main and subordinate clause, possibly connected by "that":

\noindent (1) Explicit usage, as in "do you agree that his suggestion sounds better?" More specifically, we identified sentences where SCONJ "that" immediately follows the main verb, as with the main verb "agree" in the example above. A set of rules was devised for identifying relevant sentences, filtering out cases where "that" was used in roles other than SCONJ, such as \textit{demonstrative determiner} ("I have never been to that part of the city"), \textit{demonstrative pronoun} ("that is a beautiful view"), or \textit{relative pronoun}, ("Ann is on the team that lost.").\footnote{Due to its much lower frequency, we leave the investigation of "that" as a \textit{relative conjunction} to future work.}

\noindent (2) Implicit usage, as in "my brother thinks [that] partners should always choose the former alternative", where SCONJ "that" could have been used but was deliberately omitted. The set of rules used for identifying these sentences is identical to the rules used for detection of explicit usages, except that we required the absence of "that" in the appropriate syntactic role. Appendix A.1 provides details on syntactic analysis and rules used to extract relevant sentences. Table~\ref{tbl:dataset} reports the details of the collected dataset.

\begin{table}[hbt]
\centering
\resizebox{\columnwidth}{!}{
\begin{tabular}{l|r|r}
type & sentences & mean sent. len \\ \hline
\texttt{explicit} "that" SCONJ & 40,786 & 21.85 \\
\texttt{implicit} "that" SCONJ & 57,845 & 18.07 \\
other "that" usages & 51,802 & 19.57 \\
%total & ... & ... \\
\end{tabular}
}
%\vspace{-0.1in}
\caption{Dataset details: out of over 487K sentences, almost 150K contain "that" in various syntactic roles. Note the slightly higher mean sentence length in sentences with \texttt{explicit} "that" SCONJ compared to \texttt{implicit}. We return to this observation in Section~\ref{sec:methodology}.}
\label{tbl:dataset}
\end{table}

\paragraph{Evaluation}
A random subset of 500 sentences split equally between explicit and implicit "that" usages was selected for manual evaluation by one of the authors of this paper. The evaluator was guided to check whether omitting "that" in explicit SCONJ cases would result in equally valid, meaning-preserving utterance, and vise versa -- whether adding explicit "that" in places it was omitted, would not hurt the sentence fluency and semantics. 96.4\% of the first sentence set were found valid, and 95.7\% of the second sentence set. Invalid cases include mainly ungrammatical utterances and sentences in languages other than English.

\subsection{Data Analysis}

\begin{table*}[hbt]
\centering
\resizebox{\textwidth}{!}{
\begin{tabular}{c|l}
explicit & sentence \\ \hline
\cmark & so the people of such places are easily fooled by the extremists and \textit{think} \texttt{\color{blue}{that}} polio vaccine is dangerous \\
\xmark & Well, I initially \textit{thought} \texttt{\color{blue}{[that]}} it seemed somewhat credible with a large volume of sources, and while ... \\ \hline
\cmark & Have you \textit{forgotten} \texttt{\color{blue}{that}} republicans openly \textit{admitted} \texttt{\color{blue}{that}} their \#1 priority was giving him a fight ... ? \\
\xmark & Christ, I keep \textit{forgetting} \texttt{\color{blue}{[that]}} you guys don't have the right to speak broadly of revolution. \\
%\xmark & I \textit{believe} [\texttt{that}] I would follow up with a letter of complaint both to the psychiatrist and to the governing agency. \\
%\xmark & I wouldn't \textit{say} [\texttt{that}] Lloris is the main man at fault here. \\
\end{tabular}
}
%\vspace{-0.1in}
\caption{Example sentences from the dataset with two verb lemmas -- "think" and "forget", with explicit and implicit (in square brackets) "that" usage. The main verb is in \textit{italic} and (explicit or implicit) SCONJ appears in \color{blue}{blue}.}
\label{tbl:example-sentences}
\end{table*}

We next tokenized and lemmatized the sentences using the the \href{https://spacy.io/}{spacy} python package. Table~\ref{tbl:example-sentences} presents example sentences, taken verbatim from our dataset, with explicit and implicit usages of "that" conjunction. Note that sentences with the same verb lemma (e.g., "forget") show syntactic reduction in some cases but not in others. 

\begin{figure}[h!]
\centering
\resizebox{1.0\columnwidth}{!}{
\includegraphics{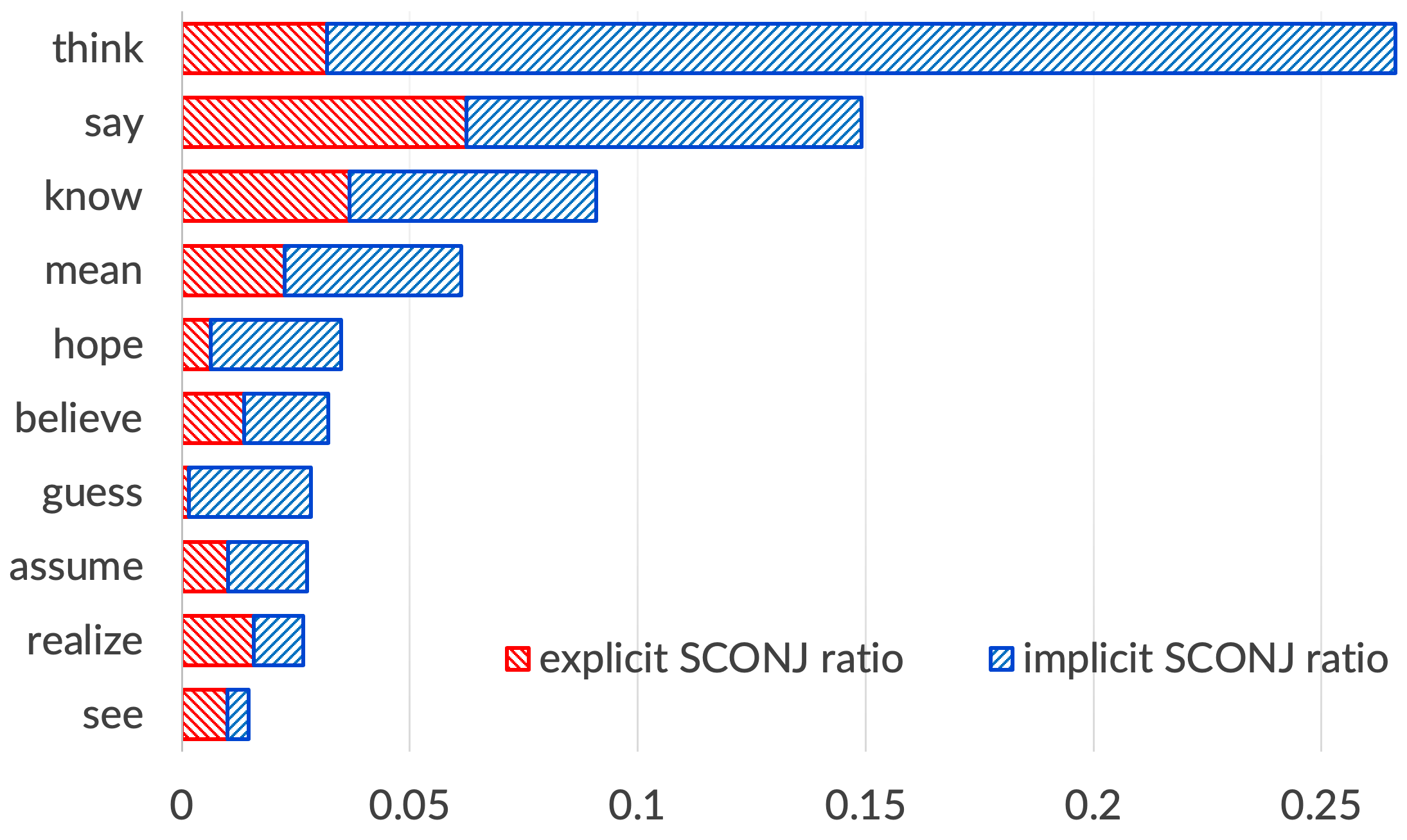}
}
%\vspace{-0.1in}
\caption{Top-10 most frequent lemmas in the data; a bar height denotes the relative ratio out of the total, and each bar is split by the relative usage of explicit and implicit "that" SCONJ. Sentences with the top-10 lemmas account for 64.7\% of all sentences in the dataset.}
\label{fig:lemma-frequency}
\end{figure}

Studying "that" omission in native and learner English, \citet{olohan2000reporting} found that the optional usage of "that" conjunction typically follows \textit{reporting} main verbs -- such as "say", "think", "suggest". Our data largely supports this observation: while the total of 434 distinct main verb lemmas were found to precede the optional "that", roughly two thirds (64.7\%) of all usages (or potential usages -- omissions) are covered by the top-10 most frequent lemmas in the dataset. Additionally, different verbs exhibit different distribution of explicit and implicit usages: while "that" is omitted in the majority of cases following lemmas "think" and "guess", other lemmas, like "say", "know", "believe", and "realize" show more balanced behavior. Figure~\ref{fig:lemma-frequency} presents the relative frequency of the top-10 most common lemmas in the dataset (bar height), and the split between explicit and implicit "that" SCONJ usages immediately following those main verbs.
In particular, the findings in Figure~\ref{fig:lemma-frequency} imply that the lemma alone does not carry sufficient predictive power about the potential syntactic reduction in subsequent subordinate clause.

\section{Methodology}
\label{sec:methodology}

%\ella{casting the task as a classification scenario, defining and extracting predictors; examples}

We define a set of factors that we were found to affect syntactic reduction choices \citep{jaeger2006speakers, jaeger2010redundancy}, and further study the magnitude of their predictive power by casting the usecase as a classification scenario. %Driven by the ample size of our dataset and the availability of contemporary LLMs, we define and calculate several novel predictors, demonstrating their predictive power on the classification problem at hand, thereby shedding new and interesting light on the usage of "that" SCONJ.
We harness the power of contemporary LLMs for reliable computation of SC onset surprisal, as well as for computation of its complementary predictor: SC onset entropy. We define the following predictors:

\paragraph{Main clause (MC) length} Previous work suggested that the conjunction is likely to be spelled out explicitly in longer sentences; in particular after a longer main clause. This predictor is computed by the number of tokens preceding the (explicit or implicit) SCONJ. As an example, in the sentence "Do you realize [that] I've never actually seen him at the office?", MC length will be assigned 3.

\paragraph{Subordinate clause (SC) length} Similar intuition suggests that the length of a subordinate clause (and more generally, the rest of the sentence) can be used as another predictor. In the example sentence above, SC length will be assigned 9.

\paragraph{Main verb frequency} \citet{jaeger2010redundancy} found negative correlation between the main clause verb frequency and the tendency to spell out "that" SCONJ. We compute the frequency of main verbs in all sentences as their relative count in the entire corpus of over 480K sentences (see Section~\ref{sec:dataset}).

\paragraph{SC subject distance} This predictor is defined as the number of words at the SC onset up to and including the SC subject. Multiple studies found positive correlation of this factor with the tendency to spell out SCONJ \citep{hawkins2001categories, hawkins2004efficiency, jaeger2010redundancy}. We extract the SC subject using the \textit{nsubj} annotation assigned by \href{https://spacy.io/}{spacy}'s dependency parser to the subordinate clause subject.

\paragraph{SC onset information density (ID)} \citet{jaeger2006speakers} and \citet{jaeger2010redundancy} computed this factor by using the simplest possible estimation, where the information of the SC onset is only conditioned on the main verb, and is operationalized by the notion of \textit{surprisal}: --\textmyfont{log \textit{p(SC onset | main verb)}}. All counts (and probabilities) were calculated from the dataset at hand. Harnessing the power of modern pretrained LLMs, we define this predictor as the probability of SC onset, conditioned on entire main clause, namely --\textmyfont{log \textit{p(SC onset | MC)}}.

Notably, \citet{jaeger2006speakers} trained the bigram model in a controlled setting where all "that" conjunctions had been omitted. Without this control, results may be circular, e.g., in cases where "that" is explicitly spelled out, the computation --\textmyfont{log \textit{p(SC onset | MC)}} could be self-evident because "that" is normally inserted between MC and SC onset (recall that SC onset denotes the opening word of the subordinate clause, "that" excluded). Since training a language model from scratch on corpora with omitted SCs is often impractical, we marginalize out the presence of "that", re-defining the SC onset surprisal to be:
\\

\noindent
\resizebox{\columnwidth}{!}{
--\textmyfont{log $\Bigl($ \textit{p(SC onset | MC)} + \textit{p(SC onset | MC $\circ$ "that")$\Bigr)$ }}
}
\\

This refined definition of SC onset surprisal eliminates the need to re-train a language model on a corpus where the SC "that" had been omitted.

%Motivated by the \textit{information density} intuition, we define two additional predictor variables, modeling the \textit{information density} hypothesis: \textit{normalized RC predictability} and \textit{RC head entropy}. 

%\paragraph{Normalized RC predictability} We hypothesize that the (presumably lower) predictability of the RC may be manifested in a more holistic manner, when considering RC as a whole, rather than its firs token. While reliable computation of a phrase predictability, operationalized by computation of its probability, is unfeasible using bi- and tri-gram LMs trained on a limited amount of data (see...), contemporary LLMs introduce an inherently natural way for this computation. We make use of the ... LLM, and calculate normalized RC predictability by averaging over RC's tokens probability, considering gradually increasing prefix.

\begin{table*}[h!]
\centering
\resizebox{0.87\textwidth}{!}{
\begin{tabular}{l|rrrr|rrrr}
predictor & \multicolumn{4}{c|}{all MC main verb lemmas} & \multicolumn{4}{c}{"think" MC main verb lemma} \\ \hline
& $\beta$ & [0.025 & 0.975] & pval sig. & $\beta$ & [0.025 & 0.975] & pval sig. \\ \hline
%& $\hat{\beta}$ & [0.025 & 0.975] & p-value & $\hat{\beta}$ & [0.025 & 0.975] & p-value \\ \hline
\texttt{const} & -0.383 & -0.41 & -0.35 & *** & -2.159 & -2.25 & -2.07 & *** \\
\texttt{MC length (tokens)} & 0.302 & 0.28 & 0.33 & *** & 0.242 & 0.17 & 0.32 & *** \\
\texttt{MC verb frequency} & -0.043 & -0.07 & -0.02 & ** & --- & --- & --- & --- \\
\texttt{SC length (tokens)} & 0.197 & 0.17 & 0.22 & *** & 0.196 & 0.12 & 0.27 & *** \\
\texttt{SC subject distance} & 0.036 & 0.01 & 0.06 & ** & 0.031 & -0.03 & 0.09 & \\ \hdashline
%\texttt{SC subject frequency} & 0.059 & & & & 0.007 & & &  \\ \hline
\texttt{SC onset surprisal} & 0.301 & 0.27 & 0.32 & *** & 0.458 & 0.38 & 0.54 & *** \\
%\texttt{SC full norm. predictability} & 0.028 & & & & 0.117 & & &  \\
\texttt{SC onset entropy} & 0.432 & 0.41 & 0.46 & *** & 0.232 & 0.15 & 0.32 & *** \\
%\texttt{MC verb tag -- VBD} & & & & & & & &  \\
%\texttt{MC verb tag -- VBG} & & & & & & & &  \\
%\texttt{MC verb tag -- VBN} & & & & & & & &  \\
%\texttt{MC verb tag -- VBP} & & & & & & & &  \\
%\texttt{MC verb tag -- VBZ} & & & & & & & &  \\

\end{tabular}
}
\vspace{-0.08in}
\caption{Logistic regression summary. $\beta$ coefficients of the scaled features mirror the sign and the relative predictor importance. 95\% CIs and p-values are reported, where "***" denotes $pval{<}0.001$ and "**" denotes $pval{<}0.01$. The MC verb frequency predictor is irrelevant in the single-main-verb-lemma experimental scenario.}
\label{tbl:regression-summary}
\end{table*}

\paragraph{SC onset entropy} We argue that the information density of the subordinate clause onset can be extended by the complementary notion of \textit{entropy} -- the expected value of the surprisal across all possible SC onsets: $H(p){=}{-}\Sigma_i{p_i*\textmyfont{log}(p_i)}$; for a given main clause MC, $p_i{=}p({w_i | MC})$, where $w_i$ is the $i^{th}$ word in the model's vocabulary $\mathcal V$. For a certain sentence prefix, entropy calculation involves the computation of the probability distribution over the model's vocabulary $\mathcal V$ for next word prediction. While the computation is practically impossible with a small corpus and an N-gram LM, this information is easily obtainable from pretrained LLMs. Although conceptually related, SC onset \textit{entropy} and SC onset \textit{surprisal} were found to be uncorrelated in our dataset: Pearson's $r$ of -0.02 was found between these two predictors.

\paragraph{Other predictors} Among additional factors investigated in prior studies are (1) SC onset frequency, (2) SC subject frequency, (3) the distance of the main verb from the SC onset, and (4) SC ambiguity ("garden path"). The first two factors were found to moderately correlate with SC onset surprisal (Pearson's $r$=-0.57) in our experiments, and hence omitted from the predictor set -- not a surprising finding given that in 84.5\% of cases SC onset is also the SC subject. The third predictor turns irrelevant in our experimental setup, where SC immediately follows the main verb. Finally, and most notably, \citet{jaeger2010redundancy} manually annotated their sentence set for SC ambiguity ("garden path"), and found this factor non-predictive of "that" omission; we, therefore, refrain from using this predictor here due to the manual effort required for "garden path" annotation in our ample data.

\section{Experimental Results and Discussion}
\label{sec:experiments}

\paragraph{Experimental Setup} We use the OPT-125m autoregressive pretrained transformer model \citep{zhang2022opt}, roughly matching the performance and sizes of the GPT-3 class of models, for computation of SC onset surprisal and entropy.  Given a sentence prefix, we first extract next token logits and convert them to a probability distribution over the lexicon by applying the softmax function. SC onset surprisal was computed by applying the natural log on the SC onset token probability given the relevant sentence prefix. SC onset entropy was computed by applying the entropy equation (see Section~\ref{sec:methodology}) on the outcome probability distribution.\footnote{Experiments with larger OPT models and decoder models from additional model families resulted in similar findings, while less efficient (higher latency). We, therefore, adhere to our choice of advanced, yet relatively small, model.}

Estimating the contextual surprisal (or entropy) per word with decoder LLMs operating at the subword level is hard; we, therefore, approximate these metrics by computing the surprisal (or entropy) over the subwords. \citet{pimentel2023effect} show that this is practically equivalent to computing a lower bound on the true contextual measurements.

Finally, logistic regression is used as a predictive model due to its effectiveness and intrepretability.

\begin{figure*}[h!]
\centering
\resizebox{1.0\textwidth}{!}{
\includegraphics{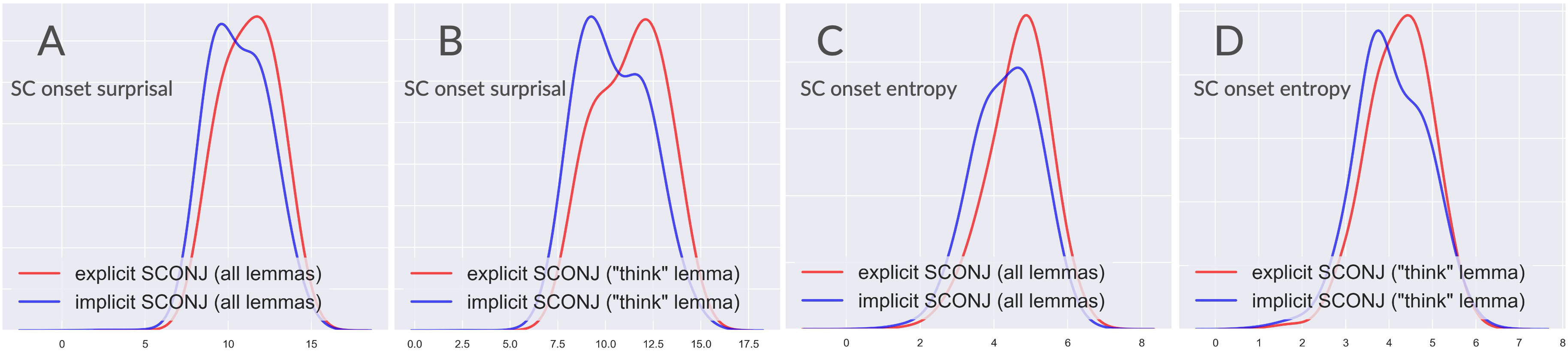}
}
\vspace{-0.25in}
\caption{Kernel density estimation plots: SC onset surprisal for explicit and implicit "that" usages, using the full lemma set (A) and the "think" lemma (B). SC onset entropy for explicit and implicit "that" usages, for the full lemma set (C) and "think" main verb lemma only (D).}
\label{fig:kernel-density}
\end{figure*}

\paragraph{Experimental Results} Our main results are presented in Table~\ref{tbl:regression-summary}. We report two scenarios: (1) all main verb lemmas preceding the SC are considered, and (2) only sentences with the most-frequent "think" main verb lemma are considered. Using these two different experimental setups, we test whether observations evident for the full set of main verbs, also emerge in a single main verb scenario. All predictors are standard-scaled for comparative analysis. The effectiveness of our predictors is supported by the considerable (in particular, much higher than chance) classification accuracy in both cases: 0.63 when using all main verbs, and 0.88 when using the "think" verb lemma only. 

\paragraph{Analysis and Discussion} Several observations emerge from the table: inline with prior studies, sentence length -- manifested in both MC and SC -- has significant positive effect on the explicit usage of "that" connecting the two clauses. One of the highest (absolute value) coefficients is assigned to SC onset \textit{surprisal}, confirming the findings by \citet{jaeger2010redundancy}. The UID hypothesis is further strengthened by the high (the highest in the all MC verb lemmas case) coefficient assigned to SC onset \textit{entropy}; that is, SC onset (non-)predictability can be viewed in a more holistic manner, where both the low predictability of the specific SC onset and the high entropy of the potential sentence continuation, carry over complementary and uncorrelated predictive power on syntactic reduction decision. %Note that the two positive coefficients (SC onset surprisal, SC onset entropy) mirror the fact that explicit usage of "that" is associated with low predictability and high entropy of "what's expected to come next". \ella{talk about SC subject distance and MC verb frequency if space permits...}
The overall picture remains consistent in the scenario where the single lemma "think" is considered (albeit SC subject distance shows insignificant), implying the robustness of our findings.

Our main findings are further strengthened by the illustration in Figure~\ref{fig:kernel-density}. Kernel density estimation of SC onset surprisal with explicit "that" usages is shifted to the right (A), reflecting the lower predictability of SC onset in this cases compared to those where "that" was omitted. This observation stays sound when only "think" main verb is considered for experiments (B). Sub-figures C and D depict the complementary entropy plots -- higher SC onset entropy in explicit "that" usages is mirrored by the right shift of the red line in both full main verb set and "think"-only cases.

The definition of surprisal inherently implies the correlation of SC onset surprisal with its frequency. Indeed, these two factors exhibit moderate negative correlation for both all lemma set and "think" lemma only (Pearson's $r$ of -0.57 and -0.47, respectively). Replacing SC onset surprisal with its frequency resulted in a slightly weaker regression model in our case, suggesting that surprisal introduces additional predictive power beyond frequency. While surprisal and frequency are highly correlated, they are typically associated with different psycholinguistic behaviours, and we leave a more thorough investigation for future work.

\section{Conclusions}
\label{sec:conclusions}
We study the UID hypothesis manifestation in syntactic reduction using a large, diverse and carefully compiled corpus of English sentences with explicit or implicit "that" subordinate conjunction. Harnessing the power of contemporary pretrained LLMs, we show that SC onset surprisal and entropy are the main factors affecting a speaker's choice to spell out the optional conjunction "that".

Last but not least, a large body of linguistic literature has studied the conditions under which complementizers (like "that" subordinate conjunction) can or cannot be omitted (inter alia \citet{erteschik1997dynamics, ambridge2008island}). We believe that future work in this field should better engage with this literature, incorporating insights for more linguistically-informed approach to the task of syntactic reduction analysis.

%\newpage
\section{Ethical Considerations}
\label{sec:ethical}

We use publicly available data to study the manifestation of UID in syntactic reduction. The use of publicly available data from social media platforms, such as Reddit, may raise normative and ethical concerns. These concerns are extensively studied by the research community as reported in e.g., \citet{proferes2021studying}. 
Here we address two main concerns. (1) Anonymity: Data used for this research can only be associated with participants' user IDs, which, in turn, cannot be linked to any identifiable information, or used to infer any personal or demographic trait. (2) Consent: \citet{jagfeld2021understanding} debated the need to obtain informed consent for using social media data mainly because it is not straightforward to determine if posts pertain to a public or private context. Ethical guidelines for social media research \citep{benton2017ethical} and practice in comparable research projects \citep{ahmed2017using}, as well as \href{https://www.redditinc.com/policies/user-agreement-september-12-2021}{Reddit's terms of use}, regard it as acceptable to waive explicit consent if users’ anonymity is protected.

We did not make use of AI-assisted technologies while writing this paper. We also did not hire human annotators at any stage of the research.
\section{Limitations}
\label{sec:limitations}

We believe that the main limitation of this work is the relatively restrictive experimental setup of sentences used to study UID principles in syntactis reduction. As an example, additional syntactic setting of interest includes sentences where "that" is used as a relative conjunction, as in "the book [that] I read last week made me quite sad...". Due to its much lower frequency in our data, we leave the investigation of "that" omission before a relative clause to future work.

The current study also limits its set of main clauses to those where the SCONJ immediately follows MC verb, not considering cases like "My boyfriend has mentioned several times [that] we should approach this guy with the offer", where the main verb "mentioned" is separated from the SC onset "we" by the "several times" phrase. However, we have reasons to believe that similar findings would be evident in these scenarios, and plan to extend the research to those cases as well.

\section*{Acknowledgements}
We are grateful to Shuly Wintner for much advice during the early stages of this work. We are also thankful to Alon Rabinovich for his help with the annotation effort for this study.

\bibliography{custom}
\bibliographystyle{acl_natbib}

\newpage
\appendix
\onecolumn
\section{Appendices}
\label{sec:appendices}

\subsection{Identification of Sentences with Optional "that" Subordinate Conjunction}
\begin{figure*}[h!]
\centering
\resizebox{0.65\textwidth}{!}{
\includegraphics{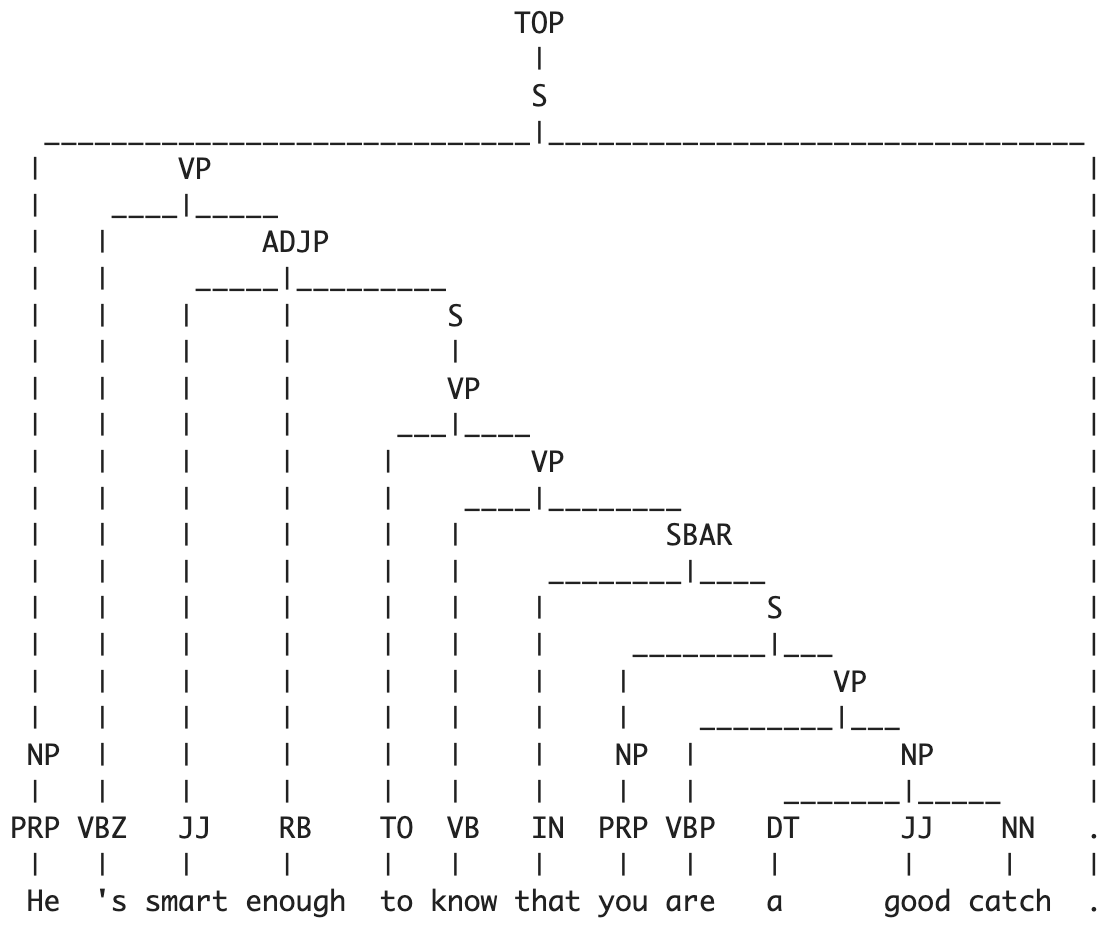}
}
%\vspace{-0.1in}
\caption{Constituency parse tree of the sentence "He's smart enough to know that you are a good catch.". Note the main verb "know" followed by the explicit SCONJ "that" and subordinate clause "you are a good catch".}
\label{fig:explicit-that}
\end{figure*}

\begin{figure*}[h!]
\centering
\resizebox{1.0\textwidth}{!}{
\includegraphics{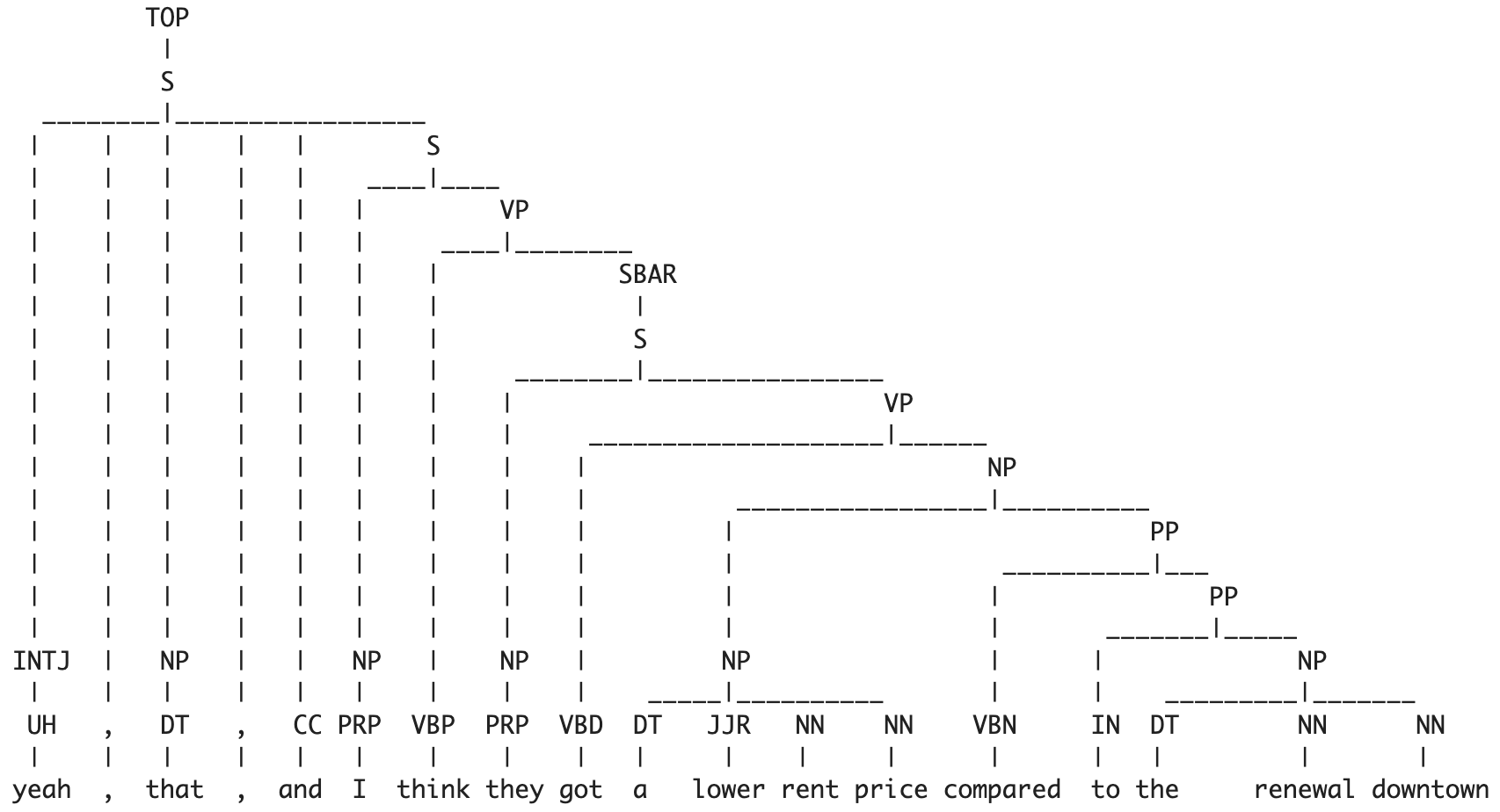}
}
%\vspace{-0.1in}
\caption{Constituency parse tree of the sentence "yeah, that, and I think they got a lower rent price compared to the renewal downtown". Note the main verb "think" followed by the omitted SCONJ "that" and subordinate clause "they got a lower rent price compared to the renewal downtown".}
\label{fig:implicit-that}
\end{figure*}

Figures~\ref{fig:explicit-that} and \ref{fig:implicit-that} depict two parsing trees of sentences with explicit and implicit usage of "that" SCONJ, respectively. After parsing a sentence, a set of rules was applied for identification of cases where "that" is used (or could have been used) in the role of subordinate conjunction connecting main and subordinate clause. As mentioned in Section~\ref{sec:dataset}, the extraction process was tuned for accurate (over 95\%) performance.

\end{document}